\documentclass[runningheads]{llncs}
\usepackage{graphicx}
\usepackage{amsmath}

\usepackage[utf8]{inputenc}
\usepackage{graphicx}
\usepackage{hyperref}
\usepackage{amsmath}
\usepackage{amsfonts}
\usepackage{latexsym}
\usepackage{epsfig}
\usepackage{multicol}
\usepackage{subfig}
\usepackage{algorithm}
\usepackage{algpseudocode}

\begin{document}
\title{Data-Driven Vehicle Trajectory Forecasting}
%\author{Shayan Jawed\inst{1}}
\author{Shayan Jawed \and
Eya Boumaiza\and
Josif Grabocka\and
Lars Schmidt-Thieme}
\institute{Information Systems and Machine Learning Lab \\ University of Hildesheim\\ Samelsonplatz 22, Hildesheim 31141, Germany \\
\email{\{shayan,eya,josif,schmidt-thieme\}@ismll.de}}
\maketitle              % typeset the header of the contribution
\begin{abstract}
An active area of research is to increase the safety of self-driving vehicles. Although safety cannot be guarenteed completely, the capability of a vehicle to predict the future trajectories of its surrounding vehicles could help ensure this notion of safety to a greater deal. We cast the trajectory forecast problem in a multi-time step forecasting problem and develop a Convolutional neural network based approach to learn from trajectory sequences generated from completely raw dataset in real-time. Results show improvement over baselines. 
  
\end{abstract}
\section{Introduction}
Safety is of paramount importance while desiging self-driving cars. It is an active area of research and never easy to answer "how safe is safe enough". The capability to predict the future trajectory of surrounding vehicles and react accordingly helps ensuring this notion of safety to a great deal. Besides safety, infering the future behaviours of surrounding vehicles in a dynamic and sufficiently complicated traffic environment grants autonomous vehicles the ability to make tactical driving decisions such as overtaking and lane changing. Also, it mimics the inherent human ability of constantly predicting every other vehicle’s movement. \\ However, given a sufficiently complex traffic environment it can be a challenging task. The difficulty lies in the fact that many factors influence a vehicle’s trajectory which can eventually be very noisy to model. These factors could include geometric structure of the road, vehicle sizes,  driver intentions etc. \\Broadly speaking, in order to design such a practical system, two steps are required. Firstly, detection and subsequent tracking of surrouding vehicles in real-time and secondly, design and implementation of less computationally complicated algorithms in order to guarantee real-time inference. 

In this paper, we study the problem of vehicle trajectory forecasting as a multi-step time series regression. While considerable progress has been made with aggressively exploring Recurrent Neural Networks and more specifically LSTM based architectures to solve similar problems, in this paper we take a different approach and instead propose a Convolutional Neural Network architecture to infer from temporal data as an alternative.

In our proposed system, we base the first step on state-of-the-art methods in object detection and tracking. Thus, the system is able to generate trajectory sequences from raw camera input ignoring the need of complex sensors such as LiDAR. Next, these trajectory sequences are fed to a Convolutional Neural Network (CNN) which is able to forecast the paths of the surrounding vehicles. Moreoever, we conduct experiments to show the outperformance of the proposed approach over standard baselines.

The rest of this paper is organized as follows. In Section II, we briefly review the related work. In Section III, we describe the dataset used and the system setup used for developing the proposed prediction framework. In Section IV, we describe the details on the proposed vehicle trajectory prediction framework. In Section V, the experimental results are provided and the paper is concluded in Section VI.

\section{Related Work}
Several recent studies have addressed the problem of vehicle trajectory forecasting.  A survey of previous work could be found in \cite{lefevre2014survey}. Previous approaches can be grouped into two major classes. The first class of approaches treat the trajectory forecasting as a time series classification task where the aim is to assign class labels such as turn left, right or stay in lane for the trajectories. For such classification, Markov based models \cite{streubel2014prediction} and Support Vector Machines have been adopted \cite{kumar2013learning}. Moreover, Neural Networks were also used in this direction where a probabilistic multilayer perceptron based approach is proposed in \cite{yoon2016multilayer} to model how likely a vehicle is to follow a particular trajectory or a lane for a given input of vehicle position history. Furthermore, the Long short term memory (LSTM) network was also investigated in \cite{khosroshahi2016surround} where a framework for activity classification of on-road vehicles using 3D trajectory cues was proposed. \cite{kim2017probabilistic} also follows a classification objective where an LSTM model is trained with sequence data to produce softmax probabilities of future vehicle position on the grid. 

On the other hand, the second class of algorithms cast the trajectory forecasting problem as a time series regression task where the aim is to predict the exact position of the car in future timesteps. Wheareas the classification based approaches make modelling the problem inherently less complex, they are less expressive than their regression based counterparts which try to predict for some time in future the path in it's entirety instead of classifying the vehicle behavior. This complete path can then be used for a more effective motion planning. More specifically, the path prediction in terms of regression was studied with various approaches such as linear regression \cite{mccullagh1989generalized}, Kalman filters\cite{kalman1960new} and non-linear Gaussian processes\cite{wang2008gaussian}. Moreover, approaches in this direction also include time-series analysis\cite{priestley1981spectral} and autoregressive models\cite{akaike1969fitting}. In addition, analogous to previous work done for trajectory classification, a large state of literature in the time series regression direction is based on LSTM based approaches. Among these approaches, \cite{altche2017lstm} proposed an LSTM model trained to predict future set of (x,y) positions for the target vehicle.  Another interesting approach was proposed in \cite{lee2017desire} where a set of hypothetical future prediction samples are first obtained through a conditional variational autoencoder, which are subsequently ranked and refined by an RNN scoring-regression module. 

%
%Approaches that take this direction involve linear regression\cite{mccullagh1989generalized}, Kalman filters\cite{kalman1960new} and non-linear Gaussian processes\cite{wang2008gaussian}. Also worth noting are time-series analysis\cite{priestley1981spectral} and autoregressive model \cite{akaike1969fitting} which have a similar objective. 

It is worth noting  that LSTMs are considered the state-of-the-art in time series forecasting currently because of  the capability of these networks to access the entire history of previous time series values using its recurrent property. Alternatively, other authors have investigated Convolutional Neural Networks (CNN) to model temporal data taking advantage of the convolutional operation that makes the number of trainable weights smaller and thus speeds up training and prediction. This motivation was explored in recent work such as \cite{mittelman2015time} where an undecimated fully convolutional neural network based on causal filtering operations was proposed for time series modeling. This architecture  introduces a wavelet transform-like deconvolution stage, which allows for the input and output lengths to match. Also, authors in \cite{binkowski2017autoregressive} proposed  a convolutional network extension of standard autoregressive models equipped with a nonlinear weighting mechanism for forecasting financial time series. Moreover, based on an adaption of the recent deep convolutional WaveNet architecture\cite{oord2016wavenet}, \cite{borovykh2017conditional} proposed a network containing stacks of dilated convolutions that allow it to access a broad range of history when forecasting. The common intuition in these works is that by modeling time series via CNNs, representative filters for repeating patterns in the series could be learned and effectively used to forecast the future values.
  
Also, previous literature lacks an approach completely based on raw camera input. Raw camera input is the most readily accessible information in real-time as access to other driving vehicles motion model or speed is not easily possible. Moreover, previous approaches rely heavily on LiDAR based tracking ignoring the recent advances in object detection and tracking. 

In this paper, we propose a two phased dynamic approach where surround vehicles are firstly detected and then tracked for a limited time. The resulting sequences are then input to a deep Convolutional Neural Network (CNN) model which outputs trajectories for the same amount of time in future. The approach is dynamic in the sense that after a selected period of time the detection and tracking of surrounding vehicles is reinitiated to generate new trajectory sequences. This way, the surround vehicles might be the same or it could be that new vehicles might enter in the time period that follows. Accordingly, trajectory forecasts would be made for each period.

\section{Dataset}
%In this section we explain the raw dataset and the preprocessing applied to generate the final trajectory dataset. 
%\subsection{Raw Udacity Dataset}
The Udacity dataset consists of 30,000 frames recorded at 20 frames per second. The data was captured with a camera mounted to the windshield of the car while driving in Mountain View California. The dataset contains a fair amount of lightning changes, number of turns, lane merges and driving on a multi-lane divided highway. Moreover, in parts of the dataset there is quite an amount of traffic as well. These characteristics make the dataset suitable for our task as they represent real-life scenarios. 

\begin{figure}[h]

	\subfloat[]{
		\centering
		\includegraphics[width=0.25\textwidth]{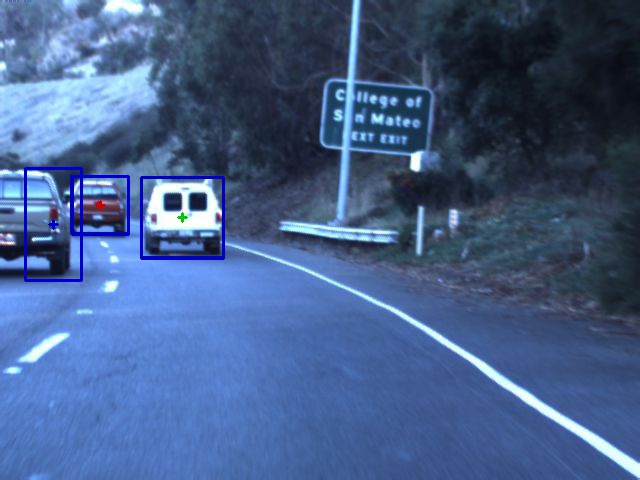}%, height=9em
	}
	\subfloat[]{
		\centering
		\includegraphics[width=0.25\textwidth]{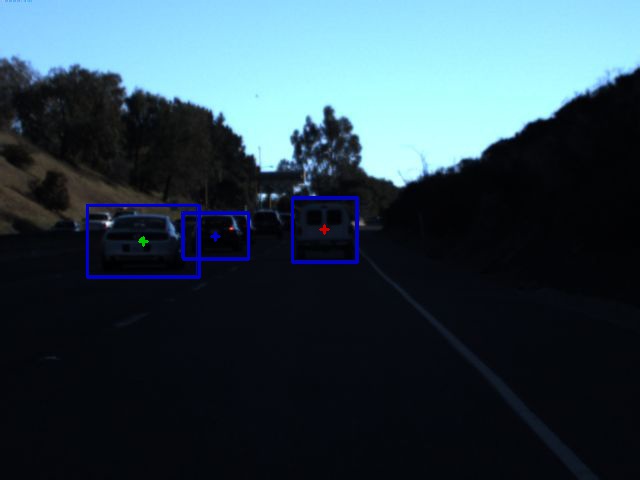}%
	}
	\subfloat[]{
		\centering
		\includegraphics[width=0.25\textwidth]{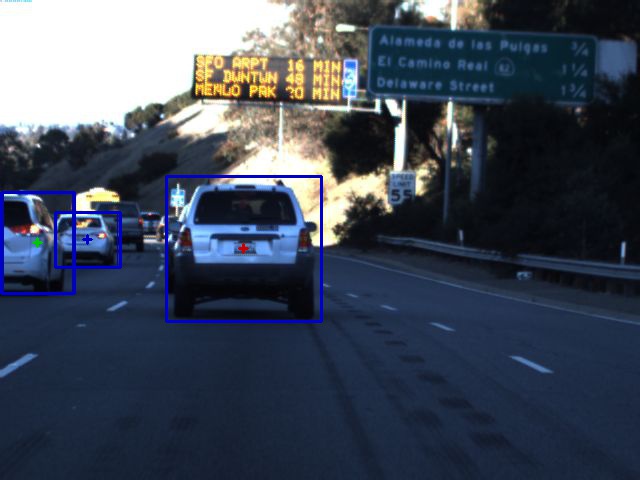}%
	}
	\subfloat[]{
		\centering
		\includegraphics[width=0.25\textwidth]{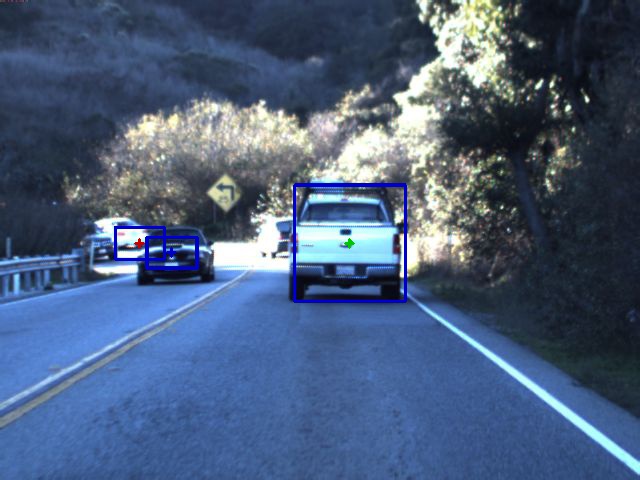}%
	}	
	\caption{Surrounding vehicles tracked}\label{fig1}
	
\end{figure}

\section{Method} 
We formulate the vehicle trajectory estimation problem as a multi-variate, multi-step time series regression process, where the objective is to predict the future trajectories of N tracked vehicles $Y=\{  Y^{(1)}, Y^{(2)}, .., Y^{(N)} \}$ given the past trajectories $X=\{ X^{(1)}, X^{(2)}, .., X^{(N)}\}$. 
The past trajectory of the  $i^{th}$ tracked vehicle  is defined as $X^{(i)} = \{(x^{(i)}_{t=0}, y^{(i)}_{t=0}), (x^{(i)}_{t=1}, y^{(i)}_{t=1}), .., (x^{(i)}_{t=\alpha}, y^{(i)}_{t=\alpha})\}$, and the future trajectory is defined  as $Y^{(i)} = \{(x^{(i)}_{t=\alpha+1}, y^{(i)}_{t=\alpha+1}), (x^{(i)}_{t=\alpha+2}, y^{(i)}_{t=\alpha+2}), .., (x^{(i)}_{t=\beta}, y^{(i)}_{t=\beta})\}$. Here,  ($x_{t}^i$,$y_{t}^i$) represents the $x$ and $y$ coordinates of the center of the $i^{th}$ surrounding vehicle in a captured frame at timestep $t$ while $\alpha$ and $\beta$ represent the maximum sequence for past and future timesteps respectively.

In this section, we describe the details of our method in the following structure: We first explain the Classic Bootstrapping method used to generate trajectory data from raw input data in Section 4.1 and then present the learning models used to predict the future trajectories in Section 4.2.

\subsection{Classic Boostrappping}
The manual labelling of vehicles in the raw image dataset for generating trajectory data is a tedious task. Instead, we follow a classical bootstrapping approach to generate trajectories for the surrounding vehicles. A first version of the system is developed to generate training data and to reduce the effort of manual labelling. The output is then corrected to insure the robustness of the bootstrapping system.

The first version of the system involves two steps, detection of surrounding vehicles $i_0$,...$i_N$ in the raw image dataset at timestep $t=0$ (first frame) and then subsequently tracking the respective vehicles until timestep $t=\beta$. This will generate the trajectories of the N detected vehicles in a time horizon of length $\beta$ with the trajectory of the $i^{th}$ vehicle being determined by the sequence of its center coordinates  $\{(x_{t=0}^i$,$y_{t=0}^i$), ($x_{t=1}^i$,$y_{t=1}^i$),...,($x_{t=\beta}^i$,$y_{t=\beta}^i$)\}.  The system is then restarted again after each $\beta$ timesteps to detect and track vehicles starting from timestep $t=k(\beta+1)$ where $k \in [0,  int(\frac{T}{\beta})]$ and T is the total number of frames in the dataset. The vehicles trajectories dataset is finally generated as the set of vehicles trajectories with time horizon length of $\beta$ for $int(\frac{T}{h})$ horizons. 

Next, the output of this system is manually corrected where a human detects the instances where the tracking failed and removes them from the dataset. For example on a curved lane or if the vehicle is detected on the opposite lane an otherwise consistent tracking might fail. An example of tracking failure could be seen in Fig 1. (d) where a vehicle traveling in the opposite direction is being tracked.  
Indeed, if a car is detected in a curved lane, it means that it is somehow far away from the driver's vehicle since it is not in the nearby area and thus it would be better not to consider it as a surrounding car. Also, it would not be relevant to consider a car in the opposite lane as a surrounding car since it will only appear for a short time compared to our tracking time horizon especially if the car is on the highway (less than a second). 

Moreover, we also develop an unsupervised approach to detect such tracking failures, as explained in further detail in the last subparagraph of this section. 

\subsubsection{Vehicle Detection using Faster R-CNN with Resnet101}
Modern object detection is based on the use of convolutional neural networks (CNNs). Object detectors such as Faster R-CNN, R-FCN, SSD have shown a lot of promise in a variety of applications including real-time detection. In order to understand how these differ in terms of speed and accuracy, authors in\cite{huang2017speed} created a framework based on Tensorflow where they vary factors such as the choice of feature extractor (e.g. VGG, Resnet etc.), image resolution etc. With respect to this analysis, and experiments conducted on the Udacity dataset, we base our vehicle detection system on the Faster R-CNN framework in\cite{ren2015faster} with the feature extractor being Resnet 101 \cite{he2016deep} pre-trained on the Microsoft COCO dataset \cite{lin2014microsoft}.

Faster R-CNN, is composed of two modules. The first module is a region proposal network (RPN) that proposes regions of an image having higher chance to contain an object. These regions are then used by the second module, Fast R-CNN \cite{girshick2015fast} for detection. 
More specifically, the entire image is firstly pushed through a feature extractor and feature maps are cropped at some intermediate layer of this extractor. The RPN module then uses these features to generate box proposals called also “anchors”. Subsequently, these anchors are then used to crop features from the same intermediate feature map and are then fed to the Fast R-CNN to predict a class and and class-specific box refinement to fit the ground truth for each anchor. Following this anchors methodology and the multi-task loss in Fast R-CNN, the loss function for Faster R-CNN for an image is given as:

\begin{equation}
L(\{p_i\}, \{t_i\}) = \frac{1}{N_{cls}} \sum_{i} L_{cls}(p_i, p_i^*)+ \lambda \frac{1}{N_{reg}} \sum_{i} p_i^* L_{reg}(t_i, t_i^*)
\end{equation} 

where $i$ is the index of an anchor and $p_i$ is the predicted probability of anchor $i$ being an 
object, vehicle in our case. The ground-truth label $p_i^*$ takes 1 if the anchor is positive, and 0 if the anchor is negative. $t_i$ is a vector representing the 4 parameterized coordinates of the predicted bounding box, and $t_i^*$ is that of the ground-truth box associated with a positive anchor.
The classification loss $L_{cls}$ is the log loss over two classes (object vs. not object) and the regression loss is  the robust loss function (smooth $L_{1}$) defined in \cite{girshick2015fast}. The term $ p_i^*L_{reg}$ means the regression loss is activated only for positive anchors ($p_i^*=1$) and is disabled otherwise ($p_i^*=0$). The outputs of the $cls$ and $reg$ layers consist of {$p_i$} and {$t_i$} respectively. The two terms are normalized by $N_{cls}$ and $N_{reg}$ and subsequently weighted by a balancing parameter $\lambda$.

%
%Modern object detection is based on the use of convolutional neural networks (CNNs). Object detectors such as Faster R-CNN, R-FCN, SSD have shown a lot of promise in a variety of applications including real-time object detection. In order to understand how these differ in terms of speed and accuracy, \cite{huang2017speed} created a framework based on Tensorflow where they vary factors such as the choice of feature extractor (e.g. VGG, Resnet etc.), image resolution etc. Based on this analysis, and experiments conducted on the Udacity dataset, we base our vehicle detection system on the framework's Faster R-CNN \cite{ren2015faster} with the feature extractor being Resnet 101 \cite{he2016deep} pretrained on the Microsoft COCO dataset \cite{lin2014microsoft}.
%Detection begins in the Faster R-CNN with the feature extractor selecting features at some intermediate level in a convolutional layer and using this to predict class-agnostic box proposals. Subsequently, these box proposals are then used to crop features from the same intermediate feature map and are then fed to the remainder of the feature extractor to predict a class and and class-specific box refinement for each proposal. 

\subsubsection{MIL Tracking}
%Vehicles detected at timestep $t_0$ are subsequently tracked for a time period of $t_1$...$t_\beta$. Tracking is based on the Online Multiple Instance Learning algorithm \cite{babenko2009visual}. Given the location of the surrounding vehicles $i_0$,...$i_n$ in the first frame instead of having multiple positive and negative patches around the location of the object which might confuse the classifier, in MIL based tracking multiple patches form a single bag and the bag is labeled positive if there exists atleast one positive example. Subsequently, a boosting classifier that maximizes the likelihood of the bags is trained. Notably, in the online algorithm the MIL classifier is updated with positive and negative example bags for every frame. 
%We use the open-source implementation of the Online Multiple Instance Learning provided with the OpenCV contrib extension \cite{opencv_library}.

Vehicles detected at timestep $t=0$ are subsequently tracked for a time period of $t=1$ ... $t=\beta$. Tracking is based on the Online Multiple Instance Learning algorithm \cite{babenko2009visual}. Given the locations of the surrounding vehicles in the first frame, instead of having multiple positive and negative patches around the location of each car which might confuse the classifier, in MIL based tracking multiple patches form a single bag and the bag of m patches $B_i = \{b_{i1}, .., b_{im}\}$ is labeled positive ($y_i =1$) if there exists at least one positive example. Subsequently, a boosting classifier is trained to maximize the likelihood of the all the bags:

\begin{equation}
log {\cal{L}} = \sum_{i} (\log p(y_i=1|B_i)) 
\end{equation} 
where  the probability of a bag being positive $p(y_i|B_i)$ is expressed in terms of its instances as follows: 

\begin{equation}
p_i(y_i=1|B_i) = 1- \prod_{j} (1-p(y_i=1|b_{ij}))
\end{equation} 

Notably, in the online algorithm the MIL classifier $H(x) = \sum_{k=1}^{K} \lambda_k h_k(x)$ is updated with positive and negative example bags for every frame by choosing the K weak classifiers sequentially as follows: 

\begin{equation}
h_k= \underset{h \in \{h_1,...,h_M\}}{\arg\max}   log {\cal{L}}(H_{k-1}+h) 
\end{equation} 
We use the open-source implementation of the Online Multiple Instance Learning provided with the OpenCV contrib extension \cite{opencv_library}.

\subsubsection{Tracking failures}
In some cases such as the one showed in Fig 1. (d), a vehicle might be detected even though it's on the other side of the highway, and subsequently tracked for a specific time. This would obviously be a tracking failure as the vehicle shall not be present in the future frames. An unsupervised approach is proposed to detect such tracking failures and to manually remove such instances from the dataframe. It works by generating the histograms of the crops of the vechicle of interest from the first few frames. It also computes the histograms of crops at all four directions (up, down, left and right) of equal length as the crop of the vehicle itself from the first frame where the detection is done. Thus the crops containing the vehicle could be thought of as positive crops and the other ones as negative. 
At the end of the tracking sequence, the histograms of crops of the vehicle from the last few frames are compared to these of the first few positive and negative using the Kullback–Leibler divergence measure (KL-divergence) defined as follows:

\begin{equation}
d(H_1,H_2)= \sum_I^N H_1(I) \log \Big( \frac{H_1(I)}{H_1(I)} \Big)
\end{equation}
If the KL-divergence is larger between the last and positive crops than the last and negatives then the system could identify it as a tracking failure. 

\subsection{Models}
%\subsubsection{Multi-Step prediction schemes}
In this section, we explain three implementation choices for multi-step prediction of a trajectory  $Y^{(i)} = \{(x^{(i)}_{t=\alpha+1}, y^{(i)}_{t=\alpha+1}), (x^{(i)}_{t=\alpha+2}, y^{(i)}_{t=\alpha+2}), .., (x^{(i)}_{t=\beta}, y^{(i)}_{t=\beta})\}$.

\subsubsection{Iterative Method} 
In the iterative method, the model predicts for a single step. This prediction is then iteratively used to predict ahead. \\To predict for one time step ahead:
\begin{equation}
Y^{(i)}_{t={\alpha+1}} = f(X^{(i)})
\end{equation} 
For two time steps ahead:
\begin{equation}
Y^{(i)}_{t={\alpha+2}} = f(X^{(i)} \cup Y_{t={\alpha+1}})
\end{equation} 
Similary the process is followed until $t={\beta}$. The technique however suffers from the problem of accumulating errors which affect future forecasts. 
\subsubsection{Joint Method}
In the joint method, a single model predicts for $t={\alpha+1}$...$t={\beta}$ all at the same time. 
\begin{equation}
Y^{(i)} = f(X^{(i)})
\end{equation}
\subsubsection{Independent Method}
Using this method we predict for multiple horizons using multiple independent models. Within each horizon there might be multiple timesteps to be predicted for which the joint strategy is used. 
For example for two n-sized horizons:
\begin{equation}
\begin{aligned}
Y^{(i)}_{t={\alpha+n}} = f(X^{(i)})   \\
Y^{(i)}_{t={(\alpha+n)+n}} = g(X^{(i)}) 
\end{aligned}
\end{equation}
Since each model is dedicated to predicting its own horizon, this method is effective where each forecast horizon is unique. \\Both, independent and joint method are better suited for problems involving forecasting for large horizons compared to the iterative strategy. Also worth noting is that while there is the appealing notion of parameter sharing accross the complete horizon to be forecasted in the joint method it is however a much more complex strategy than the other two.

\subsubsection{CNN}
A CNN architecture is built by stacking a sequence of convolutional layers. A convolutional layer is implemented with a convolution operation by sliding a filter over the input in an iterative fashion, and computing the dot products between the filter and the input. The network learns then filters that are able to recognize specific patterns. More accurately, given a one dimensional input tensor $V \in \pmb{\mathbb{R}}^{\beta}$ where  $\beta$ is the length of the trajectory sequence. We define the convolution operation, denoted by $V^{l+1}=V^{l}*W^{l}$ as follows:
\begin{equation}
\label{eq3}
\begin{split}
V^{l+1}_{i}=\sum_{i' \in \left[1,z\right]} W^{l}_{i'}*V^{l}_{ i+i'-1},  i \in \left[1,w^{l+1}\right].
\end{split}
\end{equation}Where $V^{l} \in \pmb{\mathbb{R}} ^ {\beta}$ and $V^{l+1} \in \pmb{\mathbb{R}}^{\beta}$ are the input and output tensors, respectively and $W^{l} \in \pmb{\mathbb{R}}^{z}$ is the filter to be learned. Zero padding is used to maintain the same sized output. 

Specifically in our approach each sequence is fed to the CNN to encode the sequence into a feature vector. We stack 7 convolutional layers in successive order with filter sizes 24, 32, 64, 128, 256, 512 and 1024 respectively. After the sequence has passed through each of these, we apply a global max pooling layer and three fully connected layers of dimensions 256, 128 and 64. For all layers, we use linear activations and zero-padding for same sized output. Finally once the sequence has been encoded we stack as many dense layers as needed for Joint strategy and the independent strategy with dimensions of 2 each for both x and y coordinates. The model is then trained end-to-end.  

\section{Results}
In this section we compare the proposed method with the baselines by predicting the future trajectories of the surround vehicles. We define an evaluation strategy where in order to best judge the performance of the models randomly 80\% of trajectory sequences are chosen in the training and the rest 20\% to test upon. We make a design choice and set the $\alpha=25$ and the $\beta=50$. This way the length of all vehicle trajectory sequences is fixed to be $50$. This corresponds to a horizon of approximately 2.5 seconds. By fixing the sequence length and maximum vehicles $i_{n}=3$ as such we could generate at most 1800 unique trajectory sequences from a total of 30,000 images. Out of these only 1281 trajectory sequences were non-empty given that some image sequences might have at max 3 whereas in some only 2,1 or even no surround vehicles to keep track of. Following the above defined evaluation protocol we end up with 1024 training sequences and 257 testing sequences. Given the scarcity of sequences we ignore the removal of sequences where tracking might have failed as recognized by a human or the unsupervised approach of comparing the histograms. The unsupervised approach detected a total of 580 sequences where the tracking had failed, whereas the human calculation was 553 such tracking failures. It is also worth noting that the unsupervised approach and the human results had close agreement guarenteeing the effectiveness of the unsupervised approach. 

\begin{figure}
\center
\includegraphics[width=0.5\textwidth]{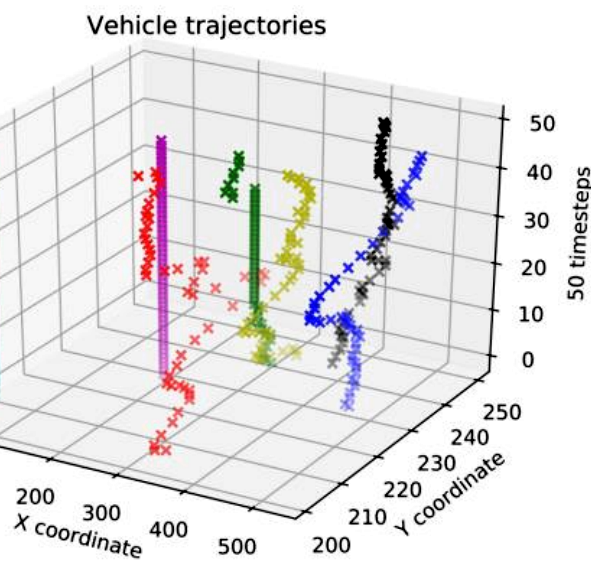}
\caption{Randomly sampled vehicle trajectories in 3D} \label{fig2}
\end{figure}

\begin{figure}[h]
	\subfloat[1]{
		\centering
		\includegraphics[width=0.5\textwidth]{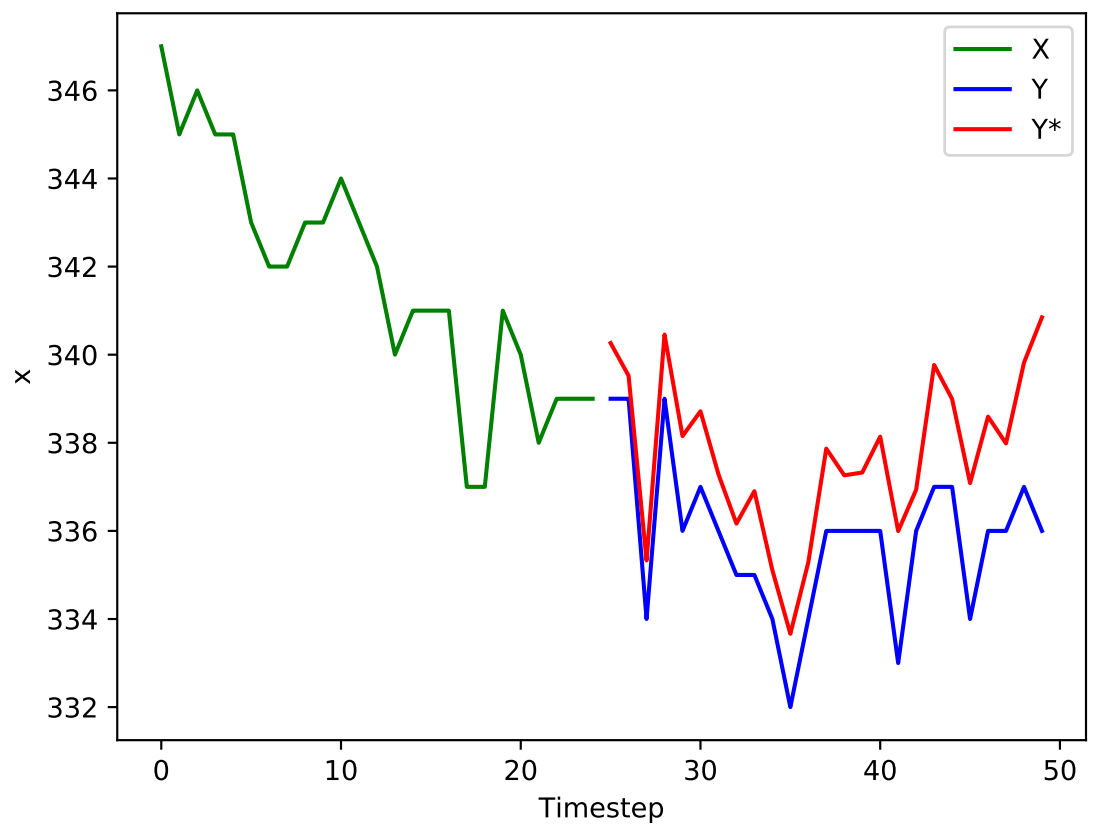}
	}
	\subfloat[2]{
		\centering
		\includegraphics[width=0.5\textwidth]{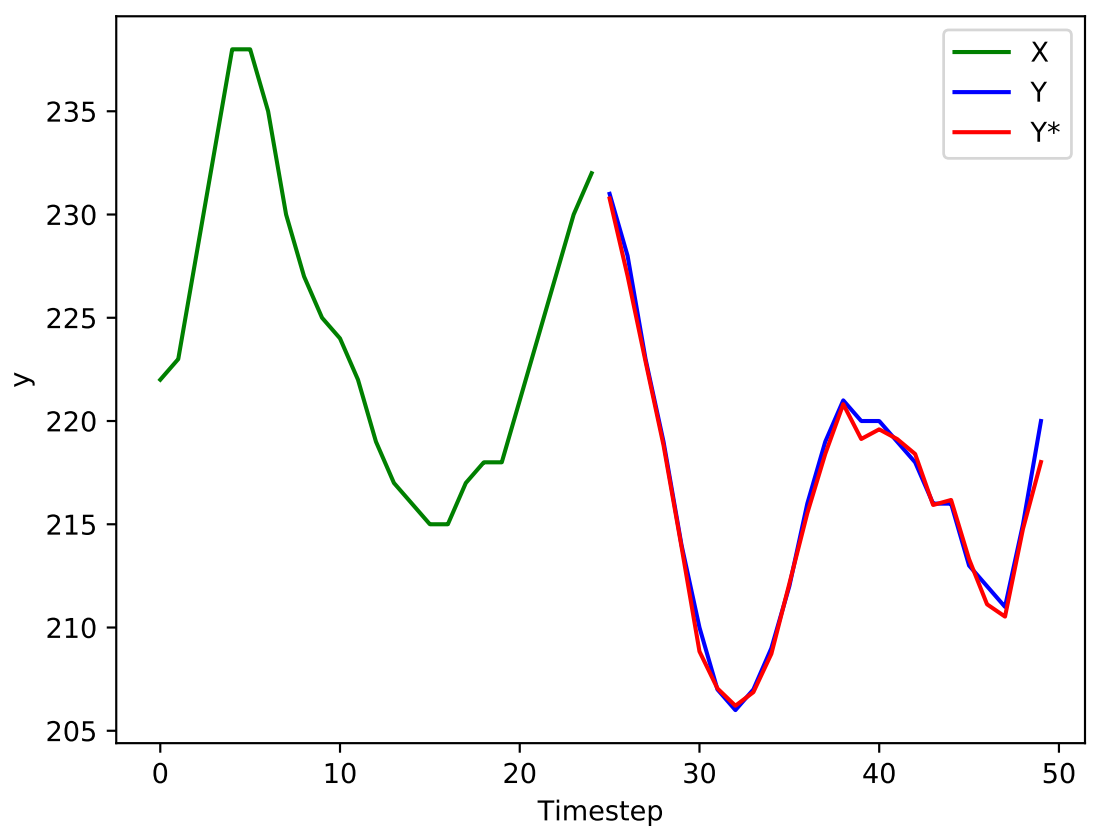}
	}
	\caption{A sampled trajectory coordinates x and y. Green, blue and red correspond to the train, test and predicted values respectively.} \label{fig3}	
\end{figure}

\begin{figure}[h]
	\subfloat[]{
		\centering
		\includegraphics[width=0.5\textwidth]{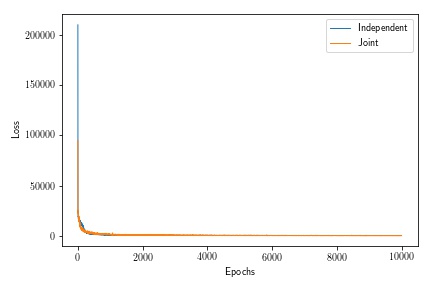}
	}
	\subfloat[]{
		\centering
		\includegraphics[width=0.54\textwidth]{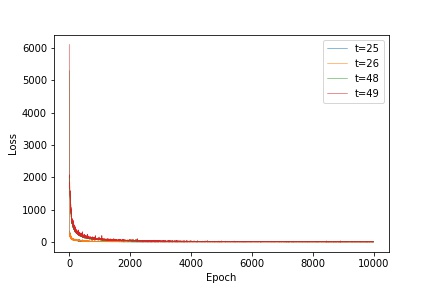}
	}
	\caption{Learning curve comparison between Joint and Independent strategies. b) Learning curves for sub-tasks of the Joint Strategy.} \label{fig4}	
\end{figure} 

\begin{figure}
		\centering
		\includegraphics[width=0.6\textwidth]{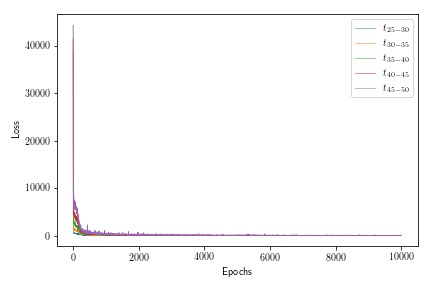}
		\caption{Learning curves for the independent methods with different horizons.} \label{fig5}
\end{figure}

\subsection{Evaluation metrics and Baselines}
We define the loss as the mean squared error between the prediction and the ground truth at multiple timesteps. It is worth noting that due to the accumulation of errors and higher inference time, we ignore training the model via the iterative approach but mentioned it before nevertheless as a possible strategy. For the independent method, 5 separate CNN models were trained each with an independent forecast horizon of 5 timesteps but with same input at $\alpha=25$. The forecast horizons and respective errors are shown in Table 1. As stated in the introduction however, the focus of the paper was on delivering real-time inference and hence the joint strategy which is the fastest among the three is adopted for all baselines and the CNN. The independent strategy for CNN is considered as baseline. Baselines include the two simple approaches of predicting the mean and the last value for all the trajectory sequences. Lastly, the CNN model is trained with joint strategy is evaluated against linear regression which estimates linear parameters by minimizing the least squared error and the Decision Tree based regression. The decision tree regression algorithm was trained by setting the maximum depth to be as much until all leaves contained 2 samples per split. 

\begin{table}[!htb]
    \begin{minipage}{.5\linewidth}
      \centering
\begin{tabular}{| c | c |}
  \hline  Independent Method Horizon & MSE \\
  \hline  $t_{25-30}$ & 558.28 \\
  \hline  $t_{30-35}$ & 1574.35 \\
  \hline  $t_{35-40}$ & 2684.40 \\
  \hline  $t_{40-45}$ & 4172.05 \\
  \hline  $t_{45-50}$ & 6192.45 \\
  \hline
 \end{tabular}
 \caption{MSE for different horizons}
    \end{minipage}%
    \begin{minipage}{.5\linewidth}
      \centering
\begin{tabular}{| c | c |}
  \hline  Method & MSE \\
  \hline  Decision Tree & 39516.815 \\
  \hline  Baseline: Last Value & 23511.61 \\
  \hline  Linear Regression & 19600.36  \\
  \hline  Baseline: Mean & 7259.255 \\
  \hline  Independent CNN Approach  & {15181.55} \\
  \hline  Joint CNN Approach & \textbf{3393.64} \\
  \hline
 \end{tabular}
 \caption{MSE for approaches}
    \end{minipage} 
\end{table}

\subsection{Learning Details}
We train the model using the Adam optimization \cite{kingma2014adam} with the initial learning rate fixed to be 0.0001. All five of the independent models and the joint model were trained for 10,000 epochs with batch size 3. The learning curves are shown in Fig. 4.

Moreover it was also observed that the losses for the timesteps close to the $\alpha$ were comparitively much less than the losses closer to $\beta$ as shown in Fig. 5. which makes sense intuitively as predicting for a much higher future horizon would be difficult.   
All models were implemented using Keras\cite{chollet2015keras} with Tensorflow background \cite{tensorflow2015-whitepaper} and trained end-to-end on Nvidia GTX 1080 Ti. Training time was approximately 28 hours each for the independent models and approximately 48 hours for the training of the joint strategy. 
\newline
\section{Conclusion}
In this work, we proposed a convolutional neural network architecture for vehicle trajectory forecasting. We approached the problem as object tracking in a sequence of images, where the goal was to forecast the position of the vehicles being tracked in future frames. Towards this end, we implemented a CNN based approach trained with the independent and joint strategies to forecast multi-time step ahead. The CNN based approach outperformed more classical methods such as linear regression, decision trees and naive baselines. 

As future work, we shall try to incorporate larger tracking sequences in order to forecast for longer periods in future. Furthermore, we shall research possible methods to output probability distributions instead of single values to yield a confidence measure for better motion planning. 
\section{Acknowledgement}
The authors gratefully acknowledge the co-funding of their work by the Ministry for Education and Research of the Federal Republic of Germany via the project ÜberDax.
\bibliography{biblio}
\bibliographystyle{splncs04}

\end{document}